\documentclass[runningheads]{llncs}
\usepackage[T1]{fontenc}
\usepackage{graphicx,verbatim}
\usepackage{amsmath} 
\usepackage{algorithm}
\usepackage{algpseudocode}

\begin{document}
\title{Decision-based AI Visual Navigation for Cardiac Ultrasounds}
%
% \begin{comment} 
%% Removed for anonymized MICCAI 2025 submission
\author{Andy Dimnaku\inst{1} \and Dominic Yurk\inst{1} \and Zhiyuan Gao\inst{1} \and 
Arun Padmanabhan\inst{2} \and Mandar Aras\inst{2} \and Yaser Abu-Mostafa\inst{1}}
\authorrunning{F. Author et al.}
\institute{California Institute of Technology, Pasadena CA 91125, USA \and
University of San Francisco, San Francisco CA 94117, USA \\
\email{adimnaku@caltech.edu}}

% \end{comment}

% \author{Anonymized Authors}  %% Added for anonymized MICCAI 2025 submission
% \authorrunning{Anonymized Author et al.}
% \institute{Anonymized Affiliations \\
%     \email{email@anonymized.com}}

\maketitle              % typeset the header of the contribution
\begin{abstract}

Ultrasound imaging of the heart (echocardiography) is widely used to diagnose cardiac diseases. However, obtaining an echocardiogram requires an expert sonographer and a high-quality ultrasound imaging device, which are generally only available in hospitals. Recently, AI-based navigation models and algorithms have been used to aid novice sonographers in acquiring the standardized cardiac views necessary to visualize potential disease pathologies. These navigation systems typically rely on directional guidance to predict the necessary rotation of the ultrasound probe. This paper demonstrates a novel AI navigation system that builds on a decision model for identifying the inferior vena cava (IVC) of the heart. The decision model is trained offline using cardiac ultrasound videos and employs binary classification to determine whether the IVC is present in a given ultrasound video. The underlying model integrates a novel localization algorithm that leverages the learned feature representations to annotate the spatial location of the IVC in real-time. Our model demonstrates strong localization performance on traditional high-quality hospital ultrasound videos, as well as impressive zero-shot performance on lower-quality ultrasound videos from a more affordable Butterfly iQ handheld ultrasound machine. This capability facilitates the expansion of ultrasound diagnostics beyond hospital settings. Currently, the guidance system is undergoing clinical trials and is available on the Butterfly iQ app.

% Authors must provide keywords and are not allowed to remove this Keyword section.

\end{abstract}

% Start of the introduction
\section{Introduction}

In the United States, approximately 6.7 million people over the age of 20 are affected by heart failure [1]. 1 in 4 people are expected to experience heart failure within their lifetime [1]. Vascular congestion is a common complication of heart failure where excess fluid builds up in blood vessels due to damaged cardiac function. If left unchecked, vascular congestion may lead to swelling of extremities, difficulty breathing, renal failure, and higher mortality [2]. Nevertheless, if detected early, it can be addressed through oral diuretics and other medications [3].

One of the most common ways to assess vascular volume is through right atrial pressure (RAP) [2]. Typically, RAP is low (0-5 mm Hg), but can elevate up to 10-30 mm Hg due to complications from heart disease. The gold standard for RAP measurements is through right heart catheterization, but it is invasive and has severe risks. Thus, a commonly applied noninvasive alternative for RAP measurement is ultrasound imaging. RAP is  measured in ultrasounds through the inferior vena cava (IVC), which is a vein connected directly to the right atrium. IVC scans are routinely collected as part of a standard transthoracic echocardiogram (TTE). Cardiologists use the IVC size to measure the RAP through the "Sniff Test". In the test, a patient takes a deep sniff and the level of contraction of the IVC walls is used to estimate the RAP. The smaller the contraction that occurs the higher the RAP.

Traditionally, a trained sonographer is needed to find the location of the IVC and a cardiologist is needed to determine the RAP. However, recently it has become viable for machine learning models to be able to diagnose RAP from a sniff video at the level of a cardiologist [4]. This allows RAP measurements to be captured without the presence of a cardiologist. However, finding the IVC on an ultrasound is a difficult task.

In order to broaden the usage of ultrasound beyond specialized sonographers, researchers have developed various instruction-based guidance systems to help novice operators obtain TTEs [5-8]. They direct the sonographer on how to orient the probe, but don't provide visual directions directly on the ultrasound image. Instead, we present an AI-Visual navigation system which guides a user to the IVC through real-time spatial location. Our system utilizes a decision-based video vision model trained offline to detect if an IVC is present. Our system uses the learned features from within the decision model for a novel localization algorithm which in real-time locates and annotates the IVC. Our system is able to detect the IVC as done in object detection without using annotated data or specifying a spatial learning goal. Furthermore, our navigation system complements traditional instruction-based guidance systems which provide movements for the ultrasound transducer (rotation, translation, etc). Our decision-based navigation system has been applied in a clinical trial within commercial state-of-the-art guidance accuracy. Specifically, we are presenting the following results:
\begin{enumerate}
    \item Our navigation system performed in a clinical trial achieving a $62\%$ accuracy without localized annotating (within state-of-the-art accuracy of commercial systems [5-6]).
    \item Our navigation system utilizes a decision model and a novel localization algorithm to annotate the IVC and guide the user in real-time. 
    \item Our navigation system had robust performance on real-world ultrasound videos of various qualities.
    \item Our localization algorithm is able to localize on the IVC zero-shot on a RAP classification model with a different architecture and learning goal.
\end{enumerate}

\noindent Our novel AI navigation system enables an operator without traditional sonograph experience to find the IVC. This has the potential to monitor heart failure in patients from their primary care facility. Additionally, our novel localization algorithm has proven potential for other guidance system tasks. The method can be easily translated to guidance systems for other veins, muscles, or organs. Our full AI-Guidance system is available on the Butterfly iQ app. To our knowledge, we are the first navigation system to use the learned feature maps of a deep learning model for real-time guidance.

\section{Methods}

Our navigation system utilizes a video-based decision model and a localization algorithm for real-time IVC guidance. The decision model is trained offline to detect if an IVC is present in the input video. Using the fully trained decision model, the localization algorithm is run utilizing features learned from within the decision model to spatially locate the IVC for real-time guidance. First, we will describe the two decision model architectures to which the localization algorithm was applied. The X3D architecture was used for the IVC navigation system run in a clinical setting [9]. The SlowFast model was used for RAP classification and provides a transfer learning example of the robustness of the localization algorithm [10]. We display both architectures in Figure 1. Subsequently, we will describe using SHAP to interpret the model's learned features [11]. Finally, we will go through the localization algorithm.

\subsection{Model Architectures}

\begin{figure}
\centerline{\includegraphics[width=1.0\columnwidth]{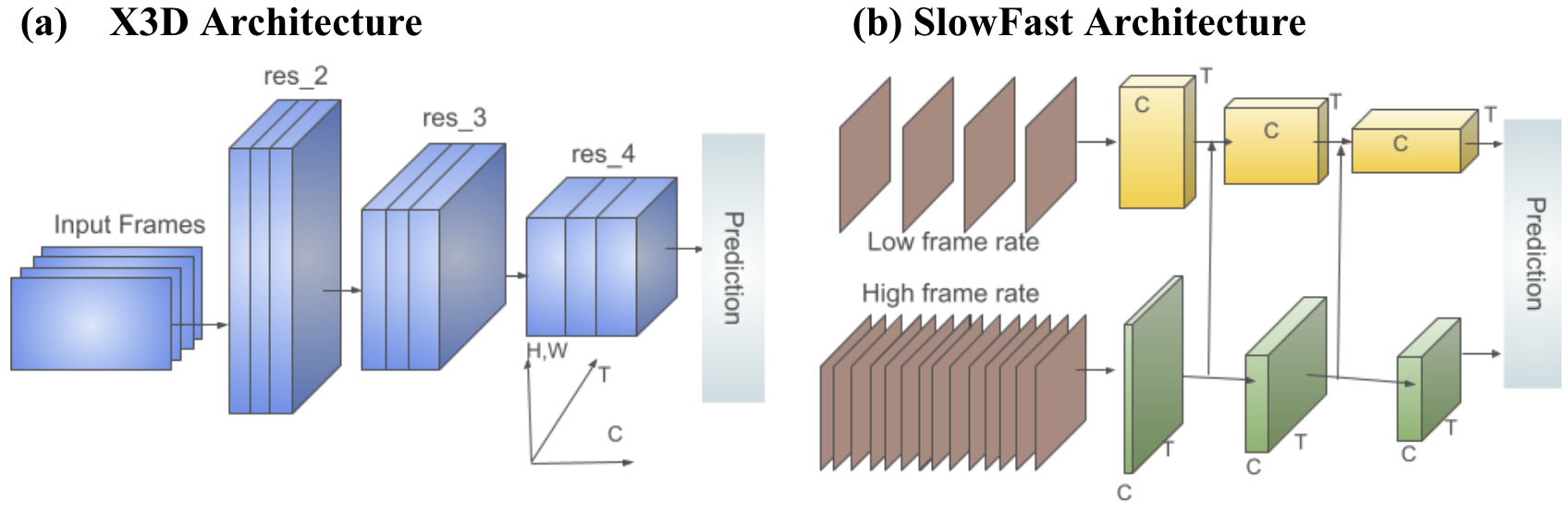}}
\caption{Figure of the X3D and SlowFast video decision model architectures. }
\end{figure}

\textbf{X3D} Our base model for IVC classification utilizes an X3D architecture [9]. The X3D architecture has high spatiotemporal understanding and is a lighter model with regard to network width and parameters. The network's lighter aspects allow for easier conversion to real-time applications. Specifically, our model is made up of 4 expanded ResNet blocks, which include several convolutions within [12].

\textbf{SlowFast} The SlowFast model takes in an input video at a high and low frame rate using a Fast and Slow pathways respectively [10]. The Fast pathway is used to efficiently learn the temporal features that change over time whereas the Slow pathway learns the distinct spatial features. We used this architecture for transfer learning on a RAP diagnostic model showing the localization algorithm's applicability across deep learning vision models.

\subsection{Model Explanation through SHAP}

Our algorithm requires understanding what information is learned by the decision model. That is, which pixels have the highest importance to the final prediction of the model. Due to the complexities of deep learning vision models, it is hard to interpret which features a model is learning for its decision. However, SHapley Additive exPlanations (SHAP) describes which portion of an input IVC video contributes the most to the final prediction [11].

SHAP breaks the input data into pieces and considers the predictive significance of each piece by evaluating different blackout combinations of the pieces on the model. SHAP uses several runs of different blackout combinations, comparing the change in the prediction to the baseline of a fully blacked-out input. The SHAP process determines the value a given portion of the input has on the prediction relative to other pieces. The results of all combinatorial blackouts are averaged together, allowing us to understand which pieces of the input are most significant to the model.

\begin{figure}
\centerline{\includegraphics[width=0.4\columnwidth]{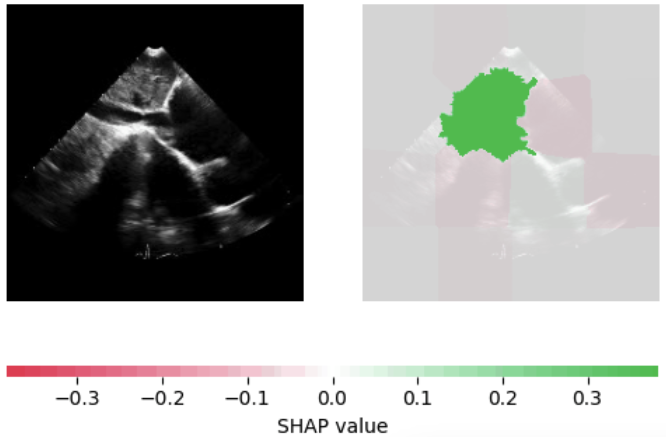}}
\caption{One frame of a 1 million iteration SHAP run of an IVC video. The SHAP values can be displayed in the below level bar signifying the relative importance. Darker color represents a larger magnitude. }
\end{figure}

In order to explain what input features our decision model was using to identify the presence or absence of the IVC we used an iterative SHAP process. We split the input IVC video into 16 pieces per frame for the 32 frames. We then ran a Monte Carlo Simulation to estimate the SHAP values. For each iteration, we generated a random permutation of blackout pieces as input into the decision model. We ran this process for 1 million iterations utilizing a different blackout combination each time. We display a sample in Figure 2. Due to the law of large numbers, we resulted in the average predictive significances of each piece. Across 30 different input videos, the SHAP process showed pieces around or including the IVC had the largest relative values.

\subsection{Localization Algorithm}

SHAP allowed us to conclude that the decision model does learn to identify localized features specific to the IVC during the classification process. However, SHAP takes multiple days to run on a GPU making it infeasible for real-time localization. We developed an algorithm utilizing the inner features of a network in order to perform SHAP-like localization in real-time. Specifically, we used the feature maps within layers of the model. These are outputs after layers where the input has been partially processed through the model. Thus, we present Algorithm 1 for real-time IVC localization.

The algorithm takes advantage of the feature map after a given layer within the vision model. It uses the learned information of the decision model for spatial location. Specifically, Algorithm 1 normalizes the channel dimension and interpolates the feature map to the original pixels. The deep layer depth of the model makes it so back-propagation for pixel correspondence or other pixel importance methods would require several magnitudes more computation, which is unrealistic for real-time use. Thus, we decided on the previous algorithm for its lighter computational efficiency.

\begin{algorithm}
\caption{IVC Navigation Localization Algorithm}\label{alg:cap}
\begin{algorithmic}
\State \textbf{Input:} video $x_i$ of shape $(1, t, h, w)$
\State \textbf{Run:} $x_i$ on X3D model resulting in prediction $p_i$
\State \textbf{If: } $p_i>0$ continue for localization otherwise stop and do not localize
\State \textbf{Extract:} feature map after the 3rd ResNet layer $y_i$ of shape $(c_y, t_y, h_y, w_y)$
\State \textbf{Normalize:} over channel dimension of $y_i$ so shape becomes $(1, t_y, h_y, w_y)$
\State \textbf{3D Spline Interpolation:} of $y_i$ to dimensions of $x_i$ $(1, t_y, h_y, w_y) \Rightarrow (1, t_x, h_x, w_x)$
\State \textbf{Store:} in set $S$ the $n$ largest valued pixels in $y_i$
\State \textbf{Remove:} from $S$ all pixels that 0 for all frames (black)
\State \textbf{Remove:} from  $S$ all pixels more than 40 pixels from the mean spatial pixel
\State \textbf{Plot:} the mean spatial pixel of $S$ with a green 20 pixel circular radius on $x_i$
\end{algorithmic}
\end{algorithm}

Algorithm 1 was run after the 3rd ResNet block layer of the X3D decision model. We decided on this feature map since it was the last layer in which the X3D architecture retained spatial-temporal information. We tested our localization algorithm across various layers of the decision model finding prior layers did not include enough information for localization whereas in later layers spatial information was lost.

\subsection{Decision Model and Localization Training}

We trained the IVC navigation model using a private dataset of transthoracic echocardiogram studies (TTE) from *****. The TTE dataset contained 4,000 TTE videos which were manually labeled as not a view of an IVC (2453 videos), a view of an IVC without a visible sniff (708 videos), and a view of an IVC with a sniff (839 videos). The decision model was trained offline separate from the localization algorithm. The localization algorithm used the trained decision model and had its parameters adjusted based on a set of 90 hospital-grade IVC scans. The localization algorithm was run only when the decision model output is positive (an IVC is present). When the decision model output is negative (no IVC) no annotating occurs. Additionally, the localized annotations results in a green 20 pixel radius circle as observed in Figures 3 and 4.

\section{Results and Discussion}

Initially, the navigation system was run using a binary decision model only displaying if an IVC was present or not without localization capabilities. This system was run in a clinical trial at ***** achieving within state-of-the-art accuracy of commercial systems in IVC guidance. The system was subsequently extended to utilize Algorithm 1 leading to further robustness. The localization algorithm was run on three different real-world datasets of various use-cases.

The navigation system was tested on high-quality hospital-grade ultrasound IVC videos. Subsequently, it was run zero-shot on the clinical trial videos mentioned. The clinical trial videos were produced by a more affordable Butterfly iQ machine allowing for testing in lower-quality settings differing in sonogram quality and shape. Finally, Algorithm 1 was run zero-shot on a RAP classification model.

\subsection{Clinical Application of Decision Navigation}

The original navigation system used only the decision output if the IVC was present without localized annotation. It was run in a clinical trial at ***** utilizing the portable and affordable Butterfly iQ ultrasound machine. The clinical trial consisted of novice sonographers with little previous ultrasound experience. As their preparation, they went through the initial Butterfly iQ tutorial lecture and were given a brief demonstration of the guidance system before starting. Testing 91 patients, our navigation system found the IVC with $62\% \pm 9.98$ accuracy, which is within state-of-the-art for IVC guidance in commercialized systems [5-6]. This provides a  backbone for our localization navigation system. This study was approved by the ***** Institutional Review Board (IRB \#*****).

\subsection{IVC Navigation with Localization}

\begin{figure}
\centerline{\includegraphics[width=1.0\columnwidth]{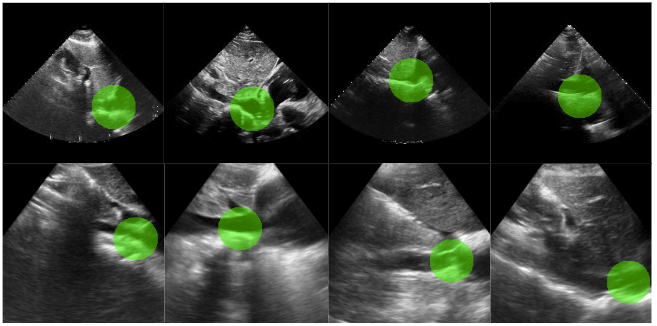}}
\caption{Samples of the navigation algorithm run on real IVC videos. The top and bottom rows are from the high quality hospital and more affortable Butterfly iQ ultrasounds respectively. The green dot is the localized location of the IVC based on Algorithm 1. }
\end{figure}

\textbf{High-quality Hospital} Given the high-quality hospital dataset, we considered a binary decision containing IVC videos with and without a sniff. The navigation was run on a separate test set of 693 videos where the decision model detected an IVC, containing both IVCs with and without sniffs. The IVC guidance system was able to localize onto the IVC in 97\% of videos on the test set. Accuracy was determined by if the 20 pixel radius annotated green circle contained the IVC by a manual reviewer.

\textbf{Butterfly iQ} The navigation system was run zero-shot on a set of low-quality IVC scans. The scans were taken from the clinical trial previously mentioned. The dataset contained 916 videos from 91 scans of patients using the Butterfly iQ where the decision model detected an IVC. The localization was able to achieve a robust performance of zero-shot $99.67\%$ localization accuracy on the IVC using the same accuracy calculations as above.

The navigation system's localization is consistently able to locate the IVC without using any spatial learning goal or annotated data. Furthermore, it was able to localize even with the large difference in the sonogram imaging shape and quality between the high-quality hospital and Butterfly iQ machines. This displays the model's capabilities working zero-shot in an out of distribution setting. Samples of the localization model are available in figure 3.

\subsection{RAP Classification model}

\begin{figure}
\centerline{\includegraphics[width=1.0\columnwidth]{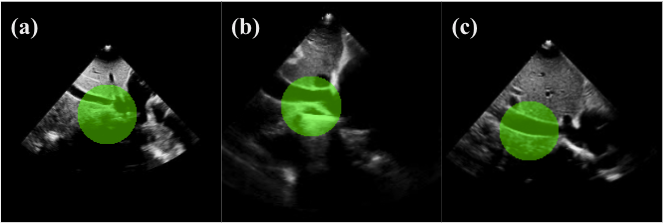}}
\caption{3 examples of Algorithm 1 run on the RAP classification model. (a), (b), and (c) are example videos at low, medium and high RAPs respectively. The images are single frames of the video for the sake of display. }
\end{figure}

We tested Algorithm 1's robustness through transfer learning by running its localization abilities zero-shot on the RAP classification model. The model was not directly trained for IVC detection as the previous IVC navigation model, but for predicting the RAP value from a given video. The learning goal was abstracted one level further through measuring how much the IVC contracts rather than if it is present. Algorithm 1 was run after the 1st block in the Slow Pathway of the SlowFast model keeping all the parameters untouched.

The model was run on a dataset of sniff scans from 30 patients from the high-quality hospital ultrasound machine. The videos need to contain the patient sniffing since that is the standard for measuring RAP through the IVC. The algorithm was run on a test set of 30 patients with 10 of low, medium, and high RAP respectively. The algorithm resulted in $100\%$ localization accuracy even with the model having a different learning objective. Examples of the RAP model localization are shown in figure 4. However, a limitation of our method is that it has not been tested for guidance of cardiac views other than the IVC.

\section{Conclusion}

The results above show the robustness of the decision-based navigation system. The sole decision-based navigation system was tested in a clinical trial providing results within the commercial state-of-the-art. The navigation system increased its robustness with IVC localization through Algorithm 1. It is able to localize the IVC in real-time on various qualities and sonogram shapes of data even in a zero-shot setting. Furthermore, the model was able to localize in an object detection manner in real-time through the use of a sole decision learning goal. The base decision model was not trained on any spatial information as in traditional object detection methods (annotated data, explicit spatial losses, etc). This suggests a new method of guidance without uncommon heavily annotated data.

We provide a novel navigation system for localized and annotated guidance in real-time for finding the IVC. Traditional navigation systems in ultrasounds tend to be instruction-based in guiding the movement of the ultrasound transducer (i.e. translation, rotation, etc). Our method is able to go hand in hand with these methods through localizing independently of instruction-based guidance. Additionally, its testing on finding the IVC shows its ability to find dynamic features.

Our navigation system through its robustness on an affordable and portable ultrasound machine has the potential to support ultrasound diagnostics beyond hospital settings. Currently, the guidance system is in clinical trials and is available on the Butterfly iQ app.


\end{document}